\pdfoutput=1

\documentclass[sigplan,10pt]{acmart}

\setcopyright{none}
\renewcommand\footnotetextcopyrightpermission[1]{}
\AtBeginDocument{%
  \providecommand\BibTeX{{%
    \normalfont B\kern-0.5em{\scshape i\kern-0.25em b}\kern-0.8em\TeX}}}

\usepackage{lipsum}
\usepackage{subcaption}
\usepackage{listings}

\usepackage{xspace}
\newcommand{\syslrn}[0]{\texttt{syslrn}\xspace}

\usepackage{soul}

\begin{document}

\title{syslrn: Learning What to Monitor for Efficient Anomaly Detection}

\author{Davide Sanvito, Giuseppe Siracusano, Sharan Santhanam, Roberto Gonzalez, Roberto Bifulco}
% \email{{name.surname}@neclab.eu}
\affiliation{%
  \institution{NEC Laboratories Europe}
}

\begin{abstract}
While monitoring system behavior to detect anomalies and failures is important, existing methods based on log-analysis can only be as good as the information contained in the logs, and other approaches that look at the OS-level software state introduce high overheads.
We tackle the problem with \syslrn, a system that first builds an understanding of a target system offline, and then tailors the online monitoring instrumentation based on the learned identifiers of normal behavior. While our \syslrn prototype is still preliminary and lacks many features, we show in a case study for the monitoring of OpenStack failures that it can outperform state-of-the-art log-analysis systems with little overhead. 
\end{abstract}

\maketitle

\pagestyle{plain}

% \vspace{-0.4cm}
\section{Introduction}
Monitoring the behavior of software to detect errors and issues is a critical task in any operational system deployment~\cite{gunawi2014bugs,gunawi2016does,datadog,dynatrace}. 
A common monitoring approach is to use automated tools to continuously collect and analyze the \textit{logs} written by the different software components~\cite{yuan2012conservative,yuan2012characterizing}. Logs contain rich information that can help to reconstruct the software execution flow, thereby enabling the detection of potential issues and errors. For instance, the reconstructed execution flow can be compared with the expected correct flow to identify the occurrence of an issue~\cite{yu2016cloudseer,nandi2016anomaly,cotroneo2020fault}.

Nonetheless, deploying log analysis systems is difficult, and their ability to detect failures is limited by which logging practice was applied during development. 
In fact, logs parsing, interpretation and correlation are all tasks that require knowledge of an application, and they have to be repeated for each of the monitored applications, and anytime log messages change due to software updates~\cite{zhang2019robust}.
Alternative monitoring approaches, such as building provenance graphs using Operating System (OS) event monitoring~\cite{wang2020you}, are instead only used in specific cases, due to the increased monitoring overhead. In fact, provenance graphs are a tool used mostly for security-critical services and for offline analysis, since: (i) Kernel-level auditing~\cite{redhat_audit} introduces significant performance overheads; (ii) the cost of updating and maintaining the graph since system boot is high and grows over time; (iii) the analysis of the (large) graph may take long time.

Our goal is to complement these existing approaches, and in some cases replace them, with an alternative that requires little domain knowledge, which is independent from software developers' practices, and which is sufficiently lightweight to be deployed in high performance scenarios. 
Towards this goal, we design \syslrn. The \syslrn's key idea is to split the monitoring system operations in two phases: during an offline training phase the monitored software behavior is observed in details to identify key \textit{indicators} of \textit{normal} behavior. During the online monitoring phase, only these indicators are continuously monitored and verified, thereby reducing the monitoring overhead.  

In this paper, we present a first minimalist implementation of a \syslrn prototype, and show that it can outperform state-of-the-art log analysis systems when monitoring a complex cloud management system like OpenStack. 

In particular, inspired by provenance graphs, in \syslrn we first track software behaviors using only information available at the interface between OS and User space applications. These interfaces are stable and have a clear semantic associated with them, thereby freeing us from the need to know application-specific semantics. Furthermore, the widespread adoption of microservice architectures makes relevant internal software events visible also at this level. Therefore, we build a complete system behavior \textit{graph} and analyze it during an offline training phase. The analysis is targeted to the identification of relevant \textit{features} that can model the \textit{normal} software behavior. While in a final version of \syslrn we envision approaches to test multiple analysis techniques in this phase, in our current prototype we introduce a simple heuristic based on \textit{bag-of-components kernels} and \textit{linear regression} algorithms. The \textit{bag-of-components} synthetically captures the structure of the software behavior graph in a vector representation. Using this representation we then build a \textit{linear regression} model to describe the relationship between the processed workload, e.g., number of service requests, and the observed graph structure. While in future \syslrn implementations we plan to complement these approaches with several other techniques, we show that this simple method is sufficient to provide a compelling anomaly detection performance in the OpenStack case study.

The analysis performed during the training phase identifies the features that characterize the software behavior, thus, during online monitoring \syslrn only collects such features, thereby tailoring the monitoring to the strictly required events that identify the system behavior. Here. we collect OS-level features relying on the recently introduced Linux's eBPF technology. eBPF allows us to inject small programs at relevant Kernel \textit{hooks}, which extract only the minimal information \syslrn requires to \textit{learn} the identified features and then perform online anomaly detection. As we will see, eBPF is efficient, which makes \syslrn's monitoring overhead 10x lower than the regular logging tasks overhead. 

We test our \syslrn prototype monitoring anomalies in OpenStack \cite{openstack}, and comparing it with DeepLog~\cite{du2017deeplog}, a state-of-the-art automated log analysis system.
We perform over 900 experiments to generate a realistic dataset instrumenting a testbed to perform common OpenStack operations, such as Virtual Machine creation, storage and network provisioning. We use the fault injection framework developed by ~\cite{cotroneo2019bad} to create failure scenarios, and in the process we collect both logs and the information required by \syslrn. Finally, we measure the ability of our partial \syslrn implementation to detect failures, and compare it with DeepLog. Our results show that the \syslrn prototype, while still limited, can outperform DeepLog in this case study. It generates a significantly lower number of false positives (Selectivity 99\%) than DeepLog (Selectivity 83\%), furthermore, \syslrn can identify a higher number of failures (Recall 98\% vs DeepLog's Recall 86\%). In fact, unlike DeepLog and log analysis systems in general, \syslrn does not depend on what software logs \cite{yuan2012characterizing}, and it is therefore better able to profile the software behavior.

These results are encouraging, and motivate us to invest further on the development of  \syslrn. Given the significant effort required to generate meaningful datasets to perform research in this area, we make available our dataset of OpenStack monitoring events. The dataset includes over 900 experiments, with per-experiment duration up to 30m, in different scenarios, and the related logs and OS-level monitoring data.

\section{Concept and Case Study}
\label{sec:concept}
Previous research on security monitoring shows that modeling the application behavior using OS-level abstractions is a powerful tool to uncover potential issues.  
In fact, a complex software application\footnote{We call \textit{application} a software system that typically runs in User space, on a single or multiple nodes, to differentiate it from OS-level software.} is typically divided into processes, with each process responsible for a subset of the application's tasks. Processes interact among them and with other processes run by the OS, or other applications, to finally achieve their goals. The type of processes, their numbers and interactions disclose relevant information about application's state and behaviour. 
However, during regular operations, systems have a potentially large number of processes, which makes their continuous system-wide monitoring expensive and inefficient. The size of the resulting monitoring data, and its complexity, is what limits security tools based on provenance graphs to offline uses.

\begin{figure}[t!]
    \centering
    \includegraphics[width=\columnwidth,bb=0 0 1399 983]{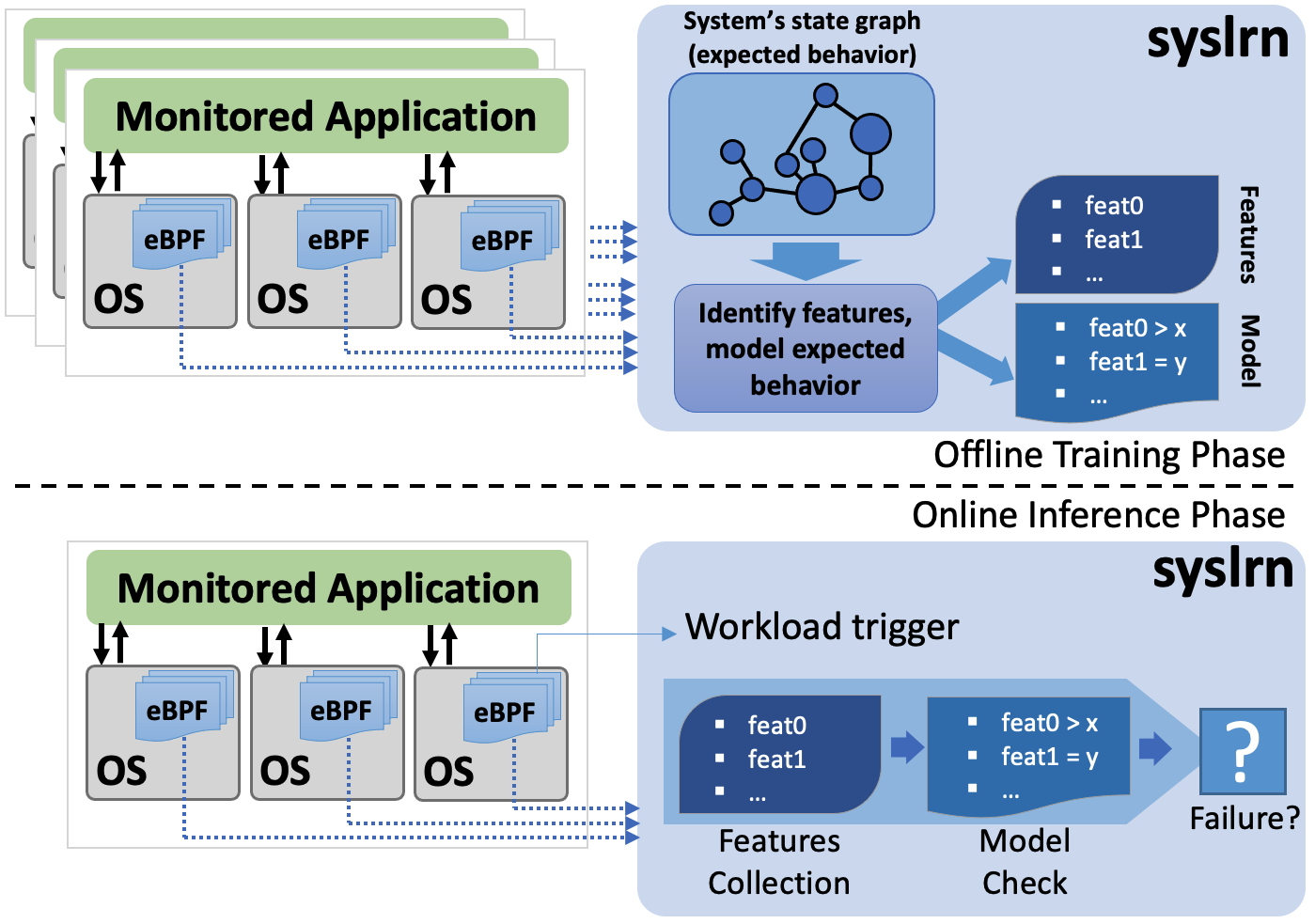}
    \caption{\syslrn overview}
    \Description[syslrn overview]{syslrn overview}
    \label{fig:arch}
    \vspace{-0.4cm}

\end{figure}

Here, we observe that, unlike security monitoring solutions that need to track the entire system state evolution over time, failure monitoring solutions might often detect a faulty behavior observing a smaller set of events. For instance, a failure may be discovered observing the lack or the unexpected presence of specific processes, and relationships among them. This moves the problem from monitoring all processes and relationships, to identifying which ones characterize a normal behavior. 
Thus, given a target application, we need to address two problems: (i) identification of the indicators of normal behavior; (ii) definition of an effective system to perform online monitoring of such indicators.

\subsection{Identifying indicators}
Identifying in a large set of processes and relationships which ones are relevant to describe the application's normal behavior is a challenging task. However, we can address it offline, which gives us the opportunity to employ more sophisticated analysis techniques.
That is, we can observe \textit{normal} system operations for a variable amount of time, and until we collect relevant information for the analysis. For instance, this can be done temporarily instrumenting the target operational system, or by running the application in a controlled environment. Then, the information collected in this way can be analyzed offline to extract the relevant indicators, during a \textit{training phase} (see Figure~\ref{fig:arch}). 

In particular, to represent the application state we resort to a graph structure, which is well-suited to capture both information about system's processes and their relationships. This finally enables the application of state-of-the-art graph analysis methods to identify common features that may serve as indicators of normal behavior. Here, the choice can be made among several methods to transform a provided graph into a \textit{vector} of features, which captures several properties of the input graph. The extracted features are then the basis on which we apply unsupervised machine learning techniques to finally derive models of the normal system behavior. 
In fact, we expect significant statistical features to emerge from the collected data, since the data we monitor is mostly related to the application control flow, and less dependent on the workload dynamics.

In summary, during the training phase \syslrn: (i) generates a graph representation of the  system state; (ii) selects and applies a technique to represent the graph in a vector of features; (iii) selects and applies an unsupervised machine learning method to model the normal application behavior. The selection of the graph vector representation influences the  unsupervised learning method that can be applied later, and both have direct influence on the type of monitoring performed during the online phase. In this paper we discuss a single method, but in future we envision \syslrn to include multiple methods that can be then selected depending on the target performance goals and application properties.

\subsection{Online monitoring}
Once the training phase provides information about what to monitor, the online monitoring phase can start. In this phase, \syslrn is deployed alongside the monitored system, and it is thus important to minimize its processing overhead.

The overhead depends on: (i) how monitoring data is collected; (ii) the type of \textit{inference} performed on the collected data. In both cases, the outcome of the training phase directly influences the operations during the online phase. 

To flexibly support different monitoring requirements, \syslrn uses eBPF. An eBPF program can be dynamically attached to different in-Kernel hooks, to intercept several system events and perform simple processing on them.
This allows us to tailor the monitoring to extract only the features identified during training.
Furthermore, \syslrn takes advantage of applications' external interfaces to drive the monitoring process. That is, applications usually expose external interfaces that trigger the execution of specific tasks, and the provisioning of the intended services. These interfaces, e.g., a web interface, can be easily monitored even with minimal knowledge about the application itself, e.g., at the OS network socket level. Thus, the overall set of processes and relationships that require monitoring can be further reduced when relating the monitoring to the reception of specific events, such as a new network connection. 

In summary, to perform online monitoring, \syslrn: (i) instruments data collection using eBPF programs tailored to the target feature extraction; and (ii) defines triggers to focus monitoring only on the relevant events of interest. 

\subsection{Case Study: OpenStack}
OpenStack is a distributed cloud infrastructure orchestrator. It comprises several modules in charge of different orchestration tasks (e.g., compute, networking, storage, identity, etc), which interact among them and with third-party software to provide their services. In this paper we focus on an OpenStack deployment that includes three main modules. First, the compute module \textit{Nova}, which is in charge of the Virtual Machine (VM) management and interacts directly with the hypervisor, to control VMs life-cycle, and with the other OpenStack components. Second, the networking module \textit{Neutron}, which is in charge of the network provisioning, and interacts with Nova and with several network functions such as virtual switches and firewalls (e.g., provided by the OS kernel).
Third, the storage module \textit{Cinder}, which is in charge of the virtual disk management. It interacts with Nova and external filesystem management services.
Overall, OpenStack is a complex distributed system with many components and inter-dependencies among them and on third-party systems. Given its central role in many cloud and telecom operator infrastructures, monitoring OpenStack and ensuring its correct operations is a task that attracted relevant research work, and which is usually used as a benchmark~\cite{yang2020far,cotroneo2020fault,ju2013fault}.

\section{\syslrn}
\label{sec:syslrn}
We now introduce our preliminary \syslrn design, with the subset of currently implemented components. We split the presentation addressing the offline training phase and online monitoring phase separately.

\subsection{Training Phase}
\begin{figure}[t!]
    \centering
    \includegraphics[width=\columnwidth,bb=0 0 1355 996]{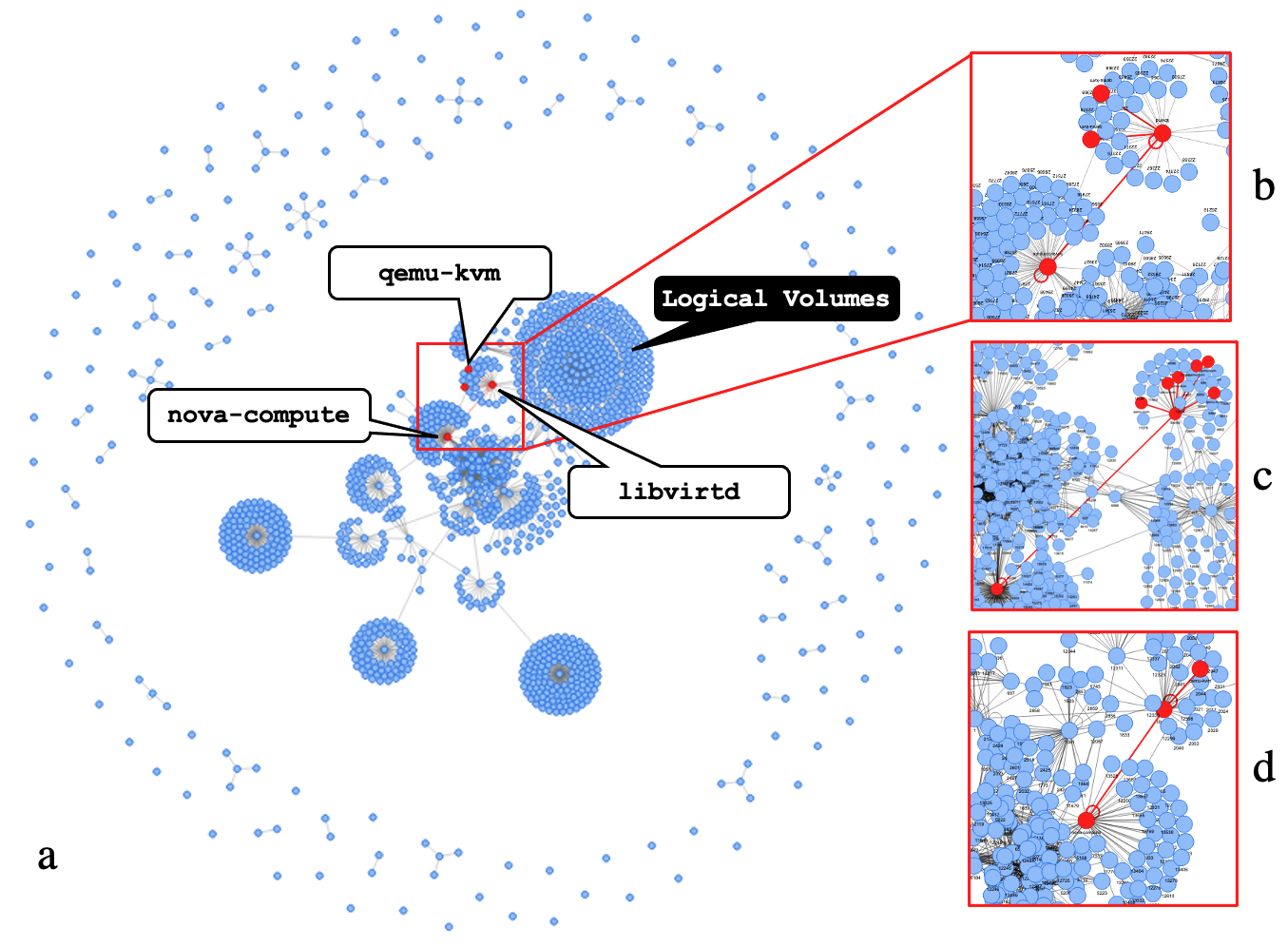}
%    \vspace{-0.4cm}
    \caption{The system graph built by \syslrn for OpenStack.}
    \Description[The system graph built by \syslrn for OpenStack.]{The system graph built by \syslrn for OpenStack.}
    \label{fig:big_graph}
    \vspace{-1em}
\end{figure}

\noindent\textbf{System graph}.
\syslrn builds a graph of the system state extracting information from the OSes running the monitored application. Such graph contains all processes (identified by their PIDs) and any interactions among them. Interactions are of three types: process creation, inter-process and network communication. Communication interactions indicate that at least one message was exchanged between two processes residing in the same, or different, hosts. That is, the graph can equally capture interactions within a single node, or across the multiple nodes the applications might run on. 

For example, Figure~\ref{fig:big_graph}a shows the system graph for an OpenStack deployment, when handling the provisioning of a single virtual machine (VM).
Peripheral nodes/components are system and background processes, while the largest connected components contain the processes related to OpenStack. 
Among these we can distinguish between OpenStack components (processes belonging to Nova, Neutron, Cinder, etc) and external services, i.e., processes not part of the OpenStack code base, but used to perform essential operations.
External processes can provide several functionalities such as: API access (e.g., \texttt{httpd}), storage (e.g., \texttt{mysqld}), and interfaces to the VM hypervisor (e.g., \texttt{libvirtd}). 

The system graph changes depending on the observation time and the workload served by the application. For instance, periodic Logical Volumes checks performed by Cinder are visible as one of the big clusters in Figure~\ref{fig:big_graph}.
The highlighted boxes show instead examples of dynamic interactions between internal and external processes in response to a service request. 
That is, when a user requests the instantiation of VMs, the request is handled by \texttt{nova-compute}, which interacts with \texttt{libvirtd} that finally communicates with the \texttt{qemu-kvm} process to create the VM. \syslrn discovers and learns about the application behavior observing such features and their evolution.
For example, when the workload is composed by a creation request for a single VM (Figure \ref{fig:big_graph}b), \texttt{libvirtd} creates two \texttt{qemu-kvm} instances
while for three VMs (Figure \ref{fig:big_graph}c), \texttt{libvirtd} creates six \texttt{qemu-kvm} instances.
Interestingly, if the VM creation fails (Figure \ref{fig:big_graph}d), only one instance of \texttt{qemu-kvm} is present.

\noindent\textbf{Feature extraction}
During training, \syslrn monitors the application in different states - i.e., before the startup, when in idle, while serving one or more workloads.
In fact, by observing the graph changes depending on the state of the application it is possible to discover: i) the system background processes; ii) the application background/maintenance processes; iii) and the processes related to workload handling, i.e., those that answer service requests.
For this last class of processes it is often important to understand their behavior in relation to the received service requests. Therefore, \syslrn monitors application's service interfaces, e.g., a socket in listening mode. This enable \syslrn to e.g., relate the amount of requests received with the changes in substructures (features) of the graph.\footnote{During training, when using synthetic workloads, this information is readily available as part of the test description.}

To reason about graph features and thereby classify software behaviors, \syslrn builds graph representations in the form of numerical vectors, called graph \textit{embeddings}~\cite{cai2018comprehensive}. Effective ways to build graph embeddings is an area of active research, therefore \syslrn can implement several approaches to address the issue, from \textit{bag-of-components}~\cite{kriege2020survey} to more advanced graph representation learning techniques based on Graph Neural Networks (GNN)~\cite{scarselli2008graph}. Focusing on the case of OpenStack, \syslrn implements a graph embedding based on \textit{bag-of-nodes}. The bag-of-nodes builds representations defining a vector with 2 dimensions for each node type (i.e., process executable names in our case): one to count the number of such nodes, and the other with the total count of their corresponding degrees (both indegree and outdegree).  

\noindent\textbf{Normal behavior model}
The embeddings are the starting point \syslrn uses to learn the \textit{normal} software behavior.
Like in the previous case, several methods can be used here, e.g., hand-tuned heuristics, clustering methods, etc. In this study, we implemented a simple heuristic that looks at the relationship between the obtained graph embeddings and the number of received service requests. Intuitively, this heuristic captures cases such as the one of the \texttt{qemu-kvm} process described earlier: both its instances counter and its degrees counter grow linearly with the number of VM requests (Fig~\ref{fig:big_graph}). 
In particular, \syslrn fits a Linear Regression (LR) model for each feature, i.e., each embedding vector's dimension, and selects among them the ones for which the LR fits well the relation with the number of processed workloads. Here, \syslrn uses the \textit{Coefficient of Determination} $R^2$ as measure of goodness-of-fit. That is, \syslrn discovers this way which features have a linear relationship with the number of received service requests.
In the case of OpenStack, only 26 features out of 152 were selected from the embeddings vector.
For example, among the selected features, some processes have an instance counter which is linear with the number of requests (e.g. \texttt{nova}, \texttt{lvcreate}), others have instead a degree counter which is linear (e.g. \texttt{ovsdb-server}) and for others both the types of counter have a linear behaviour (e.g. \texttt{brctl}, \texttt{qemu-kvm}, \texttt{iscsiadm}).
It is worth noticing that the set of selected features includes both processes associated to the three main OpenStack components as well as generic OS processes required for the VMs operations.

\subsection{Monitoring Phase}
\label{sec:sysmon}

\noindent\textbf{Features monitoring}
\syslrn performs online monitoring using a set of small eBPF programs attached to kernel probes (kprobe). These programs can be changed at runtime, effectively leading to different monitoring instrumentation configurations. 
To define these configurations, \syslrn backtracks the features selected during the training phase, mapping them to the OS primitive used to monitor them in first place. As mentioned earlier, this step is directly dependent on the graph embedding adopted during the training phase.
For instance, to collect the 26 features required for the OpenStack case, \syslrn monitors only OS' blocking stream sockets, and some process creation primitives.   
This boils down to the use of only 8 kprobe and corresponding eBPF programs. 
Furthermore, the eBPF programs record only the information needed to build the features: process name; parent PID and PID for process spawn; PIDs and network endpoints in case of communication primitives. This reduces to the minimum the overhead of executing the program each time the kprobe is invoked (i.e., the average size of the \syslrn's eBPF program is ~60 instructions). Most of the monitored functions are executed only at the process creation or when the communication channel is established, further reducing the monitoring overhead.

To perform the backtracking of the features, currently, the developer of a \syslrn training pipeline has to explicitly define the backtracking rules. We leave the automation of this step to future work.

\noindent\textbf{Anomaly detection} 
The anomaly detection module is periodically triggered to process the monitored features to check if the collected values fit the normal behavior model learned during the training phase. For our OpenStack case study, this corresponds to checking a simple ensemble of linear regression models, which verify that each feature is linearly evolving according to the number of service requests processed by the system.\footnote{As mentioned earlier, monitoring the service interface is a way to estimate the number of requests at runtime.}

\section{Evaluation}

We evaluate here \syslrn failure detection capabilities, comparing it with DeepLog~\cite{du2017deeplog} a state-of-the-art log-based failure detection system. Then we perform a microbenchmark to estimate the introduced runtime overhead.

\subsection{Dataset Generation}
While there are public available datasets for log-based failure detection, to the best of our knowledge none of them records both OS-level events and logs. Thus, we generated a new dataset\footnote{\syslrn dataset is available at \cite{syslrn_dataset}.} that records both application logs and OS-level events during OpenStack (\textit{Pike} version) normal and failure runs.
To record failures, we extended the failure injection framework presented in~\cite{cotroneo2019bad}, e.g., to support the execution of multiple concurrent requests.
With this setup, we performed 935 experiments injecting a single failure point in one OpenStack component (Nova, Neutron or Cinder), while running one or more homogeneous workloads composed by all the operations needed to create, start, stop and delete a VM.
The final dataset contains 190 failure-free experiments and 745 experiments where a failure was injected, with each experiment lasting at most 30 minutes.

We used the generated dataset to train both \syslrn and DeepLog.
It should be noted that the former only uses OS-level events, while the latter only uses application logs. In our dataset we record both of them in each experiment to enable a fair comparison of the performance of the two approaches.
Both systems use unsupervised mechanisms, thus we train only on failure-free data.
\footnote{We share the same hypothesis of DeepLog: labeled anomalous data are hard to obtain and anomalies not in the training data might be missed by supervised methods.}
Detection performance is instead evaluated using both failure-free and fail-run experiments.
We perform 10-fold cross validation to split the failure-free experiments between train/test sets. The test set includes always all the fail-runs (which are not used for training, due to the unsupervised training strategy).

\subsection{Failure detection}
\begin{figure}[t!]
    \centering
    \includegraphics[width=\columnwidth, bb=0 0 512 199]{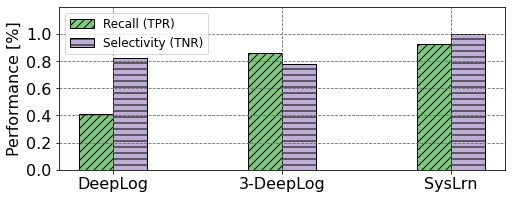}
    \vspace{-1em}
    \caption{Failure detection performance for: DeepLog in its original configuration; 3 combined DeepLog instances, each monitoring a different OpenStack component; and \syslrn}
    \Description[Failure detection performance for: DeepLog in its original configuration; 3 combined DeepLog instances, each monitoring a different OpenStack component; and \syslrn]{Failure detection performance for: DeepLog in its original configuration; 3 combined DeepLog instances, each monitoring a different OpenStack component; and \syslrn}
    \label{fig:deeplog_vs_syslrn}
    \vspace{-1em}
\end{figure}

\noindent\textbf{Baselines} DeepLog \cite{du2017deeplog} is a log-based anomaly detection system which is based on a Long Short-Term Memory (LSTM) model trained on sequences of log messages. We compare \syslrn against two different baselines. In the first one, a single DeepLog model is used to evaluate sequences of logs (identified by the instance-id) coming from a single OpenStack component (Nova). This is the default configuration of DeepLog when monitoring OpenStack~\cite{du2017deeplog}, and we verified that on the original OpenStack dataset presented in \cite{openstack_dataset} we could get similar detection performance. However, failures experienced by other components (Neutron and Cinder) are not necessarily reflected into Nova logs. Indeed, this is a typical limitation of log-based systems, which need to perform monitoring of logs belonging to different components in custom ways. 
We therefore designed a second baseline to overcome this limitation, using a dedicated DeepLog model per each OpenStack component, and combining their results. Here, it should be noted that it is not possible to combine all components' logs into a single logs stream, since anyway DeepLog requires a way to relate logs belonging to a common execution flow together. To do so, DeepLog uses rules based on identifiers mined within the logs. However, Nova, Neutron and Cinder have no obvious identifier among their logs that could be used to link logs from one component with those of the others. 

\noindent\textbf{Metrics}
We measure the True Positive Rate (Recall) and the True Negative Rate (Selectivity). The former tells how well the system identifies the failures, whereas the latter tells how well the system identifies non-failures. We do not report other commonly-used metrics, such as Precision, since they are misleading when considering highly imbalanced data, like in our case~\cite{van2022log}. 
Figure~\ref{fig:deeplog_vs_syslrn} shows average results across the 10 cross-validation splits: \syslrn outperforms both baselines for both metrics. The following subsections provide additional details.

\subsubsection{Single DeepLog}
When using log sequences extracted from Nova, a single DeepLog instance cannot detect 59\% of the failures, and wrongly classifies 17\% of the non failure cases (False Positives). This poor performance is caused by the inability to detect Neutron and Cinder failures from Nova logs, and it is a general problem for any log-based anomaly detection system. In fact, DeepLog can detect failures when critical errors appear in the logs (e.g., \textit{[instance <inst\_id>] Instance failed to spawn}) or when logs contain incorrect event sequences (e.g., \textit{[instance <inst\_id>] Instance destroyed successfully}, followed by \textit{[instance <inst\_id>] VM Resumed }). 
However, (i) not all error conditions are logged with IDs required by DeepLog to build execution flow, and (ii) not all errors cause an incorrect sequence of events. Moreover, relevant events happening in the other components are not considered for the detection. For instance, Nova logs when a volume is attached to a VM (\textit{[instance <inst\_id>] Attaching volume <vol\_id> to dev...}), but does not log any information on the correct functioning of the volume itself, which is instead contained inside Cinder logs.
That is, some failures cannot be detected monitoring only Nova logs.

\subsubsection{Combined DeepLog}
When using 3 DeepLog instances, for each component we use a different identifier to correlate logs in a flow: instance-id, network-id and volume-id, for Nova, Neutron and Cinder, respectively. The system declares an experiment as anomalous whenever at least one of the models detects an anomaly. In this case Recall jumps up to 86\%, but we also observe a small reduction in Selectivity (78\%), since false positives from each model are combined. Despite the significant effort we invested in trying to improve DeepLog performance in this case, we could not identify a better strategy to detect failure cases using multiple components' logs. Indeed, some error-related events may appear in the logs across components, but this information is hard to extract since such logs report different sequence ids (e.g., instance-id in Nova, and network-id in Neutron logs). Relating these logs would require extensive use of expert knowledge, to change DeepLog's log template definitions or to instrument directly OpenStack to carry the identifier across components.

\subsubsection{\syslrn}
\syslrn achieves 98\% Recall, meaning that only 2\% of fail runs are undetected (false negatives), and 99\% Selectivity, i.e., 1\% of failure-free experiments are misclassified as failures (false positives).
\syslrn has significantly better detection performance since: (i) it naturally relates failures happening to different components, thanks to the graph structure used to learn the application behavior; (ii) it detects failures that do not have any effect on logs but that do affect other processes in the system.
For instance in some failures of the VM disk creation routines, logs do not show any error, yet \syslrn detects an unexpectedly high number of \texttt{lvcreate} process instances.  Likewise, some failures in the creation of Neutron's virtual router interfaces are not reflected in the logs, but they affect the number of \texttt{ovs-vsctl} commands issued in the system.

\subsection{Monitoring Overhead}

\begin{table}[t!]
\setlength{\tabcolsep}{3pt}
\begin{center}
\normalsize
% \footnotesize
% \tiny
\begin{tabular}{|c|c|c|c|c|}
\hline
Operation & \vtop{\hbox{\strut Baseline}\hbox{\strut (no mon)}} & \vtop{\hbox{\strut Log-based}\hbox{\strut monitoring}}  & \vtop{\hbox{\strut eBPF program}\hbox{\strut (w/ user code)}} \\
\hline
 SET & 48.8k & 17.2k (-64.73\%) &  47.4k (-2.78\%) \\
 GET & 48.3k & 17.8k (-63.43\%) &  47.1k (-2.61\%) \\
 LPUSH & 48.6k & 17.2k (-64.51\%) &  47.4k (-2.36\%) \\
 LPOP & 49.7k & 17.1k (-65.63\%) &  48.6k (-2.19\%) \\
 \hline
\end{tabular}
\end{center}
\caption{redis-server throughput in req/s: logging vs eBPF}
\label{tab:redis-cost2}
\vspace{-2em}
\end{table}

To investigate the overhead of running OS-level feature extraction with eBPF, we run a microbenchmark using a different application as monitoring target. In fact, the OpenStack VM generation workload has a relatively low number of service requests over time, and it is unsuitable to perform a stress test with a larger number of requests per second (it would imply the creation of many VMs)~\cite{ventre2016performance,manco2017my}. 
Instead, we monitor Redis, a high performance key-value store that heavily relies on communication to perform operations like get, set and push and pop \cite{redis}.
Redis allows us to generate stress load to show more clearly the eBPF monitor overhead. 

We configured the redis-server to receive requests over a unix socket, and we used the redis-benchmark tool with 50 concurrent clients. Connection keep alive is disabled to ensure that connection operations happen on each request. 
The eBPF monitoring function is invoked once per each client request. We also enable/disable logging on the server side to measure the overhead of logs collection related to the requests.
Table~\ref{tab:redis-cost2} shows the results. Activating eBPF monitoring introduces a 3\% drop in the requests per second handled by the system. Enabling system logging reduces performance by up to 65\%, an over 10x higher overhead.
In most cases logs are required, however, for some performance critical deployments \syslrn may provide a more efficient monitoring alternative.

\section{Discussion}
\label{sec:discussion}

\syslrn uses a graph representation of the OS-level events to learn the normal software behavior during an offline training phase, and then uses this knowledge to tailor the monitoring instrumentation and reduce the online monitoring overhead. In a case study focused on failure detection for OpenStack, \syslrn outperforms state-of-the-art log-based anomaly detection systems, both in Recall and Selectivity. Here a key advantage is the \syslrn ability to relate failures happening across different software components, and the ability to reconstruct system state beyond what is reported in the logs.

However, we presented only a preliminary prototype of \syslrn, which includes a minimal subset of functionality, and our evaluation and deployment models are still preliminary.
In first place, \syslrn was demonstrated in a single application case study, and using a simplified subset of potential workloads. While this was enough to show the advantages of the approach when compared to other state-of-the-art solutions, a more thorough evaluation is required to properly understand \syslrn benefits and limitations.
We plan to test our approach on a larger set of applications, focusing on the ones based on microservices.
Furthermore, in this paper we did not address issues such as the timing of the features collection and anomaly detection. In our current tests we make simplifying assumptions, performing anomaly detection leveraging information about the expected completion time of a service request. While we believe these issues can be all addressed, we have not explored them in depth.
Finally, in future we plan to extend \syslrn with multiple graph representation and normal behavior modeling methods, and to automate the selection among those based on the measured performance. We release the datasets used to build the results in this paper to enable the community to add and evaluate additional anomaly detection methods.

\section*{Acknowledgements}
This work has received funding from the European Union’s Horizon 2020 research and innovation programme under grant agreement No. 101017171 (``MARSAL'') and No. 883335 (``PALANTIR''). This paper reflects only the authors’ views and the European Commission is not responsible for any use that may be made of the information it contains.

\bibliographystyle{ACM-Reference-Format}
\bibliography{biblio}


\begin{thebibliography}{26}

%%% the Web address.

\ifx \showCODEN    \undefined \def \showCODEN     #1{\unskip}     \fi
\ifx \showDOI      \undefined \def \showDOI       #1{#1}\fi
\ifx \showISBNx    \undefined \def \showISBNx     #1{\unskip}     \fi
\ifx \showISBNxiii \undefined \def \showISBNxiii  #1{\unskip}     \fi
\ifx \showISSN     \undefined \def \showISSN      #1{\unskip}     \fi
\ifx \showLCCN     \undefined \def \showLCCN      #1{\unskip}     \fi
\ifx \shownote     \undefined \def \shownote      #1{#1}          \fi
\ifx \showarticletitle \undefined \def \showarticletitle #1{#1}   \fi
\ifx \showURL      \undefined \def \showURL       {\relax}        \fi
% invisible to TeX
\providecommand\bibfield[2]{#2}
\providecommand\bibinfo[2]{#2}
\providecommand\natexlab[1]{#1}
\providecommand\showeprint[2][]{arXiv:#2}

\bibitem[\protect\citeauthoryear{Cai, Zheng, and Chang}{Cai
  et~al\mbox{.}}{2018}]%
        {cai2018comprehensive}
\bibfield{author}{\bibinfo{person}{Hongyun Cai}, \bibinfo{person}{Vincent~W
  Zheng}, {and} \bibinfo{person}{Kevin Chen-Chuan Chang}.}
  \bibinfo{year}{2018}\natexlab{}.
\newblock \showarticletitle{A comprehensive survey of graph embedding:
  Problems, techniques, and applications}.
\newblock \bibinfo{journal}{\emph{IEEE Transactions on Knowledge and Data
  Engineering}} \bibinfo{volume}{30}, \bibinfo{number}{9}
  (\bibinfo{year}{2018}), \bibinfo{pages}{1616--1637}.
\newblock


\bibitem[\protect\citeauthoryear{Cotroneo, De~Simone, Liguori, and
  Natella}{Cotroneo et~al\mbox{.}}{2020}]%
        {cotroneo2020fault}
\bibfield{author}{\bibinfo{person}{Domenico Cotroneo}, \bibinfo{person}{Luigi
  De~Simone}, \bibinfo{person}{Pietro Liguori}, {and} \bibinfo{person}{Roberto
  Natella}.} \bibinfo{year}{2020}\natexlab{}.
\newblock \showarticletitle{Fault injection analytics: A novel approach to
  discover failure modes in cloud-computing systems}.
\newblock \bibinfo{journal}{\emph{IEEE Transactions on Dependable and Secure
  Computing}} (\bibinfo{year}{2020}).
\newblock


\bibitem[\protect\citeauthoryear{Cotroneo, De~Simone, Liguori, Natella, and
  Bidokhti}{Cotroneo et~al\mbox{.}}{2019}]%
        {cotroneo2019bad}
\bibfield{author}{\bibinfo{person}{Domenico Cotroneo}, \bibinfo{person}{Luigi
  De~Simone}, \bibinfo{person}{Pietro Liguori}, \bibinfo{person}{Roberto
  Natella}, {and} \bibinfo{person}{Nematollah Bidokhti}.}
  \bibinfo{year}{2019}\natexlab{}.
\newblock \showarticletitle{How bad can a bug get? an empirical analysis of
  software failures in the openstack cloud computing platform}. In
  \bibinfo{booktitle}{\emph{Proceedings of the 2019 27th ACM Joint Meeting on
  European Software Engineering Conference and Symposium on the Foundations of
  Software Engineering}}. \bibinfo{pages}{200--211}.
\newblock


\bibitem[\protect\citeauthoryear{Datadog}{Datadog}{2022}]%
        {datadog}
\bibfield{author}{\bibinfo{person}{Datadog}.} \bibinfo{year}{2022}\natexlab{}.
\newblock
  \bibinfo{title}{\url{https://www.datadoghq.com/product/log-management/}}.
\newblock \bibinfo{howpublished}{Online; accessed 16-February-2022}.
\newblock


\bibitem[\protect\citeauthoryear{dataset}{dataset}{2022}]%
        {openstack_dataset}
\bibfield{author}{\bibinfo{person}{OpenStack dataset}.}
  \bibinfo{year}{2022}\natexlab{}.
\newblock
  \bibinfo{title}{\url{https://github.com/logpai/loghub/tree/master/OpenStack/}}.
\newblock \bibinfo{howpublished}{Online; accessed 16-February-2022}.
\newblock


\bibitem[\protect\citeauthoryear{Du, Li, Zheng, and Srikumar}{Du
  et~al\mbox{.}}{2017}]%
        {du2017deeplog}
\bibfield{author}{\bibinfo{person}{Min Du}, \bibinfo{person}{Feifei Li},
  \bibinfo{person}{Guineng Zheng}, {and} \bibinfo{person}{Vivek Srikumar}.}
  \bibinfo{year}{2017}\natexlab{}.
\newblock \showarticletitle{Deeplog: Anomaly detection and diagnosis from
  system logs through deep learning}. In \bibinfo{booktitle}{\emph{Proceedings
  of the 2017 ACM SIGSAC conference on computer and communications security}}.
  \bibinfo{pages}{1285--1298}.
\newblock


\bibitem[\protect\citeauthoryear{Dynatrace}{Dynatrace}{2022}]%
        {dynatrace}
\bibfield{author}{\bibinfo{person}{Dynatrace}.}
  \bibinfo{year}{2022}\natexlab{}.
\newblock
  \bibinfo{title}{\url{https://www.dynatrace.com/platform/observability/}}.
\newblock \bibinfo{howpublished}{Online; accessed 16-February-2022}.
\newblock


\bibitem[\protect\citeauthoryear{Gunawi, Hao, Leesatapornwongsa, Patana-anake,
  Do, Adityatama, Eliazar, Laksono, Lukman, Martin, et~al\mbox{.}}{Gunawi
  et~al\mbox{.}}{2014}]%
        {gunawi2014bugs}
\bibfield{author}{\bibinfo{person}{Haryadi~S Gunawi}, \bibinfo{person}{Mingzhe
  Hao}, \bibinfo{person}{Tanakorn Leesatapornwongsa}, \bibinfo{person}{Tiratat
  Patana-anake}, \bibinfo{person}{Thanh Do}, \bibinfo{person}{Jeffry
  Adityatama}, \bibinfo{person}{Kurnia~J Eliazar}, \bibinfo{person}{Agung
  Laksono}, \bibinfo{person}{Jeffrey~F Lukman}, \bibinfo{person}{Vincentius
  Martin}, {et~al\mbox{.}}} \bibinfo{year}{2014}\natexlab{}.
\newblock \showarticletitle{What bugs live in the cloud? a study of 3000+
  issues in cloud systems}. In \bibinfo{booktitle}{\emph{Proceedings of the ACM
  symposium on cloud computing}}. \bibinfo{pages}{1--14}.
\newblock


\bibitem[\protect\citeauthoryear{Gunawi, Hao, Suminto, Laksono, Satria,
  Adityatama, and Eliazar}{Gunawi et~al\mbox{.}}{2016}]%
        {gunawi2016does}
\bibfield{author}{\bibinfo{person}{Haryadi~S Gunawi}, \bibinfo{person}{Mingzhe
  Hao}, \bibinfo{person}{Riza~O Suminto}, \bibinfo{person}{Agung Laksono},
  \bibinfo{person}{Anang~D Satria}, \bibinfo{person}{Jeffry Adityatama}, {and}
  \bibinfo{person}{Kurnia~J Eliazar}.} \bibinfo{year}{2016}\natexlab{}.
\newblock \showarticletitle{Why does the cloud stop computing? lessons from
  hundreds of service outages}. In \bibinfo{booktitle}{\emph{Proceedings of the
  Seventh ACM Symposium on Cloud Computing}}. \bibinfo{pages}{1--16}.
\newblock


\bibitem[\protect\citeauthoryear{Ju, Soares, Shin, Ryu, and Da~Silva}{Ju
  et~al\mbox{.}}{2013}]%
        {ju2013fault}
\bibfield{author}{\bibinfo{person}{Xiaoen Ju}, \bibinfo{person}{Livio Soares},
  \bibinfo{person}{Kang~G Shin}, \bibinfo{person}{Kyung~Dong Ryu}, {and}
  \bibinfo{person}{Dilma Da~Silva}.} \bibinfo{year}{2013}\natexlab{}.
\newblock \showarticletitle{On fault resilience of OpenStack}. In
  \bibinfo{booktitle}{\emph{Proceedings of the 4th annual Symposium on Cloud
  Computing}}. \bibinfo{pages}{1--16}.
\newblock


\bibitem[\protect\citeauthoryear{Kriege, Johansson, and Morris}{Kriege
  et~al\mbox{.}}{2020}]%
        {kriege2020survey}
\bibfield{author}{\bibinfo{person}{Nils~M Kriege}, \bibinfo{person}{Fredrik~D
  Johansson}, {and} \bibinfo{person}{Christopher Morris}.}
  \bibinfo{year}{2020}\natexlab{}.
\newblock \showarticletitle{A survey on graph kernels}.
\newblock \bibinfo{journal}{\emph{Applied Network Science}}
  \bibinfo{volume}{5}, \bibinfo{number}{1} (\bibinfo{year}{2020}),
  \bibinfo{pages}{1--42}.
\newblock


\bibitem[\protect\citeauthoryear{Manco, Lupu, Schmidt, Mendes, Kuenzer, Sati,
  Yasukata, Raiciu, and Huici}{Manco et~al\mbox{.}}{2017}]%
        {manco2017my}
\bibfield{author}{\bibinfo{person}{Filipe Manco}, \bibinfo{person}{Costin
  Lupu}, \bibinfo{person}{Florian Schmidt}, \bibinfo{person}{Jose Mendes},
  \bibinfo{person}{Simon Kuenzer}, \bibinfo{person}{Sumit Sati},
  \bibinfo{person}{Kenichi Yasukata}, \bibinfo{person}{Costin Raiciu}, {and}
  \bibinfo{person}{Felipe Huici}.} \bibinfo{year}{2017}\natexlab{}.
\newblock \showarticletitle{My VM is Lighter (and Safer) than your Container}.
  In \bibinfo{booktitle}{\emph{Proceedings of the 26th Symposium on Operating
  Systems Principles}}. \bibinfo{pages}{218--233}.
\newblock


\bibitem[\protect\citeauthoryear{Nandi, Mandal, Atreja, Dasgupta, and
  Bhattacharya}{Nandi et~al\mbox{.}}{2016}]%
        {nandi2016anomaly}
\bibfield{author}{\bibinfo{person}{Animesh Nandi}, \bibinfo{person}{Atri
  Mandal}, \bibinfo{person}{Shubham Atreja}, \bibinfo{person}{Gargi~B
  Dasgupta}, {and} \bibinfo{person}{Subhrajit Bhattacharya}.}
  \bibinfo{year}{2016}\natexlab{}.
\newblock \showarticletitle{Anomaly detection using program control flow graph
  mining from execution logs}. In \bibinfo{booktitle}{\emph{Proceedings of the
  22nd ACM SIGKDD International Conference on Knowledge Discovery and Data
  Mining}}. \bibinfo{pages}{215--224}.
\newblock


\bibitem[\protect\citeauthoryear{OpenStack}{OpenStack}{2022}]%
        {openstack}
\bibfield{author}{\bibinfo{person}{OpenStack}.}
  \bibinfo{year}{2022}\natexlab{}.
\newblock \bibinfo{title}{\url{https://www.openstack.org/}}.
\newblock \bibinfo{howpublished}{Online; accessed 16-February-2022}.
\newblock


\bibitem[\protect\citeauthoryear{Redis}{Redis}{2022}]%
        {redis}
\bibfield{author}{\bibinfo{person}{Redis}.} \bibinfo{year}{2022}\natexlab{}.
\newblock \bibinfo{title}{\url{https://redis.io/}}.
\newblock \bibinfo{howpublished}{Online; accessed 16-February-2022}.
\newblock


\bibitem[\protect\citeauthoryear{Reference}{Reference}{2019}]%
        {redhat_audit}
\bibfield{author}{\bibinfo{person}{RHEL Audit~System Reference}.}
  \bibinfo{year}{2019}\natexlab{}.
\newblock
  \bibinfo{title}{\url{https://access.redhat.com/articles/4409591\#audit-record-types-2}}.
\newblock \bibinfo{howpublished}{Online; accessed 16-February-2022}.
\newblock


\bibitem[\protect\citeauthoryear{Scarselli, Gori, Tsoi, Hagenbuchner, and
  Monfardini}{Scarselli et~al\mbox{.}}{2008}]%
        {scarselli2008graph}
\bibfield{author}{\bibinfo{person}{Franco Scarselli}, \bibinfo{person}{Marco
  Gori}, \bibinfo{person}{Ah~Chung Tsoi}, \bibinfo{person}{Markus
  Hagenbuchner}, {and} \bibinfo{person}{Gabriele Monfardini}.}
  \bibinfo{year}{2008}\natexlab{}.
\newblock \showarticletitle{The graph neural network model}.
\newblock \bibinfo{journal}{\emph{IEEE transactions on neural networks}}
  \bibinfo{volume}{20}, \bibinfo{number}{1} (\bibinfo{year}{2008}),
  \bibinfo{pages}{61--80}.
\newblock


\bibitem[\protect\citeauthoryear{syslrn dataset}{syslrn dataset}{2022}]%
        {syslrn_dataset}
\bibfield{author}{\bibinfo{person}{syslrn dataset}.}
  \bibinfo{year}{2022}\natexlab{}.
\newblock
  \bibinfo{title}{\url{https://github.com/nec-research/syslrn-EuroMLSys22}}.
\newblock \bibinfo{howpublished}{Online}.
\newblock


\bibitem[\protect\citeauthoryear{Van~Le and Zhang}{Van~Le and Zhang}{2022}]%
        {van2022log}
\bibfield{author}{\bibinfo{person}{Hoang Van~Le} {and} \bibinfo{person}{Hongyu
  Zhang}.} \bibinfo{year}{2022}\natexlab{}.
\newblock \showarticletitle{Log-based Anomaly Detection with Deep Learning: How
  Far Are We?}. In \bibinfo{booktitle}{\emph{2022 IEEE/ACM 44th International
  Conference on Software Engineering (ICSE)}}. IEEE.
\newblock
\newblock
\shownote{to appear.}


\bibitem[\protect\citeauthoryear{Ventre, Pisa, Salsano, Siracusano, Schmidt,
  Lungaroni, and Blefari-Melazzi}{Ventre et~al\mbox{.}}{2016}]%
        {ventre2016performance}
\bibfield{author}{\bibinfo{person}{Pier~Luigi Ventre}, \bibinfo{person}{Claudio
  Pisa}, \bibinfo{person}{Stefano Salsano}, \bibinfo{person}{Giuseppe
  Siracusano}, \bibinfo{person}{Florian Schmidt}, \bibinfo{person}{Paolo
  Lungaroni}, {and} \bibinfo{person}{Nicola Blefari-Melazzi}.}
  \bibinfo{year}{2016}\natexlab{}.
\newblock \showarticletitle{Performance evaluation and tuning of virtual
  infrastructure managers for (micro) virtual network functions}. In
  \bibinfo{booktitle}{\emph{2016 IEEE Conference on Network Function
  Virtualization and Software Defined Networks (NFV-SDN)}}. IEEE,
  \bibinfo{pages}{141--147}.
\newblock


\bibitem[\protect\citeauthoryear{Wang, Hassan, Li, Jee, Yu, Zou, Rhee, Chen,
  Cheng, Gunter, et~al\mbox{.}}{Wang et~al\mbox{.}}{2020}]%
        {wang2020you}
\bibfield{author}{\bibinfo{person}{Qi Wang}, \bibinfo{person}{Wajih~Ul Hassan},
  \bibinfo{person}{Ding Li}, \bibinfo{person}{Kangkook Jee},
  \bibinfo{person}{Xiao Yu}, \bibinfo{person}{Kexuan Zou},
  \bibinfo{person}{Junghwan Rhee}, \bibinfo{person}{Zhengzhang Chen},
  \bibinfo{person}{Wei Cheng}, \bibinfo{person}{Carl~A Gunter},
  {et~al\mbox{.}}} \bibinfo{year}{2020}\natexlab{}.
\newblock \showarticletitle{You Are What You Do: Hunting Stealthy Malware via
  Data Provenance Analysis.}. In \bibinfo{booktitle}{\emph{NDSS}}.
\newblock


\bibitem[\protect\citeauthoryear{Yang, Wu, Pattabiraman, Wang, and Li}{Yang
  et~al\mbox{.}}{2020}]%
        {yang2020far}
\bibfield{author}{\bibinfo{person}{Yong Yang}, \bibinfo{person}{Yifan Wu},
  \bibinfo{person}{Karthik Pattabiraman}, \bibinfo{person}{Long Wang}, {and}
  \bibinfo{person}{Ying Li}.} \bibinfo{year}{2020}\natexlab{}.
\newblock \showarticletitle{How far have we come in detecting anomalies in
  distributed systems? an empirical study with a statement-level fault
  injection method}. In \bibinfo{booktitle}{\emph{2020 IEEE 31st International
  Symposium on Software Reliability Engineering (ISSRE)}}. IEEE,
  \bibinfo{pages}{59--69}.
\newblock


\bibitem[\protect\citeauthoryear{Yu, Joshi, Xu, Jin, Zhang, and Jiang}{Yu
  et~al\mbox{.}}{2016}]%
        {yu2016cloudseer}
\bibfield{author}{\bibinfo{person}{Xiao Yu}, \bibinfo{person}{Pallavi Joshi},
  \bibinfo{person}{Jianwu Xu}, \bibinfo{person}{Guoliang Jin},
  \bibinfo{person}{Hui Zhang}, {and} \bibinfo{person}{Guofei Jiang}.}
  \bibinfo{year}{2016}\natexlab{}.
\newblock \showarticletitle{Cloudseer: Workflow monitoring of cloud
  infrastructures via interleaved logs}.
\newblock \bibinfo{journal}{\emph{ACM SIGARCH Computer Architecture News}}
  \bibinfo{volume}{44}, \bibinfo{number}{2} (\bibinfo{year}{2016}),
  \bibinfo{pages}{489--502}.
\newblock


\bibitem[\protect\citeauthoryear{Yuan, Park, Huang, Liu, Lee, Tang, Zhou, and
  Savage}{Yuan et~al\mbox{.}}{2012b}]%
        {yuan2012conservative}
\bibfield{author}{\bibinfo{person}{Ding Yuan}, \bibinfo{person}{Soyeon Park},
  \bibinfo{person}{Peng Huang}, \bibinfo{person}{Yang Liu},
  \bibinfo{person}{Michael~M Lee}, \bibinfo{person}{Xiaoming Tang},
  \bibinfo{person}{Yuanyuan Zhou}, {and} \bibinfo{person}{Stefan Savage}.}
  \bibinfo{year}{2012}\natexlab{b}.
\newblock \showarticletitle{Be conservative: Enhancing failure diagnosis with
  proactive logging}. In \bibinfo{booktitle}{\emph{10th USENIX Symposium on
  Operating Systems Design and Implementation (OSDI 12)}}.
  \bibinfo{pages}{293--306}.
\newblock


\bibitem[\protect\citeauthoryear{Yuan, Park, and Zhou}{Yuan
  et~al\mbox{.}}{2012a}]%
        {yuan2012characterizing}
\bibfield{author}{\bibinfo{person}{Ding Yuan}, \bibinfo{person}{Soyeon Park},
  {and} \bibinfo{person}{Yuanyuan Zhou}.} \bibinfo{year}{2012}\natexlab{a}.
\newblock \showarticletitle{Characterizing logging practices in open-source
  software}. In \bibinfo{booktitle}{\emph{2012 34th International Conference on
  Software Engineering (ICSE)}}. IEEE, \bibinfo{pages}{102--112}.
\newblock


\bibitem[\protect\citeauthoryear{Zhang, Xu, Lin, Qiao, Zhang, Dang, Xie, Yang,
  Cheng, Li, et~al\mbox{.}}{Zhang et~al\mbox{.}}{2019}]%
        {zhang2019robust}
\bibfield{author}{\bibinfo{person}{Xu Zhang}, \bibinfo{person}{Yong Xu},
  \bibinfo{person}{Qingwei Lin}, \bibinfo{person}{Bo Qiao},
  \bibinfo{person}{Hongyu Zhang}, \bibinfo{person}{Yingnong Dang},
  \bibinfo{person}{Chunyu Xie}, \bibinfo{person}{Xinsheng Yang},
  \bibinfo{person}{Qian Cheng}, \bibinfo{person}{Ze Li}, {et~al\mbox{.}}}
  \bibinfo{year}{2019}\natexlab{}.
\newblock \showarticletitle{Robust log-based anomaly detection on unstable log
  data}. In \bibinfo{booktitle}{\emph{Proceedings of the 2019 27th ACM Joint
  Meeting on European Software Engineering Conference and Symposium on the
  Foundations of Software Engineering}}. \bibinfo{pages}{807--817}.
\newblock


\end{thebibliography}

\end{document}